\documentclass[nohyperref]{article}

\usepackage{microtype}
\usepackage{graphicx}
\usepackage{subfigure}
\usepackage{booktabs} 

\usepackage{hyperref}

\usepackage[accepted]{icml2022}

\usepackage{amsmath}
\usepackage{amssymb}
\usepackage{mathtools}
\usepackage{amsthm}

\DeclareMathOperator*{\argmax}{arg\,max}

\usepackage[capitalize,noabbrev]{cleveref}

\theoremstyle{plain}

\theoremstyle{definition}

\theoremstyle{remark}

\usepackage[textsize=tiny]{todonotes}

\usepackage{longtable}

\begin{document}

\twocolumn[
\icmltitle{A Comprehensive Review of Spiking Neural Networks: Interpretation, Optimization, Efficiency, and Best Practices}



\icmlsetsymbol{equal}{*}

\begin{icmlauthorlist}
\icmlauthor{Kai Malcolm}{equal,rice_ece}
\icmlauthor{Josue Casco-Rodriguez}{equal,rice_ece}
\end{icmlauthorlist}

\icmlaffiliation{rice_ece}{Department of Electrical and Computer Engineering, Rice University, Texas, United States}

\icmlcorrespondingauthor{Kai Malcolm}{kai.malcolm@rice.edu}
\icmlcorrespondingauthor{Josue Casco-Rodriguez}{josue.casco-rodriguez@rice.edu}

\icmlkeywords{Machine Learning, Spiking Neural Networks, Sparsity, Convex Optimization, Geometry, Discrete, Spike, Energy, Efficiency, Interpretable, Interpretation, Normalization, Initialization, Architecture, Accuracy}

\vskip 0.3in
]



\printAffiliationsAndNotice{\icmlEqualContribution} 

\newcommand\josue[1]{{\color{red}\sf{[Josue: #1]}}}
\newcommand\kai[1]{{\color{blue}\sf{[Kai: #1]}}}

\begin{abstract}
Biological neural networks continue to inspire breakthroughs in neural network performance. And yet, one key area of neural computation that has been under-appreciated and under-investigated is biologically plausible, energy-efficient spiking neural networks, whose potential is especially attractive for low-power, mobile, or otherwise hardware-constrained settings. We present a literature review of recent developments in the interpretation, optimization, efficiency, and accuracy of spiking neural networks. Key contributions include identification, discussion, and comparison of cutting-edge methods in spiking neural network optimization, energy-efficiency, and evaluation, starting from first principles so as to be accessible to new practitioners.
\end{abstract}

\section{Introduction}
Spiking neural networks (SNNs) offer a perspective on machine learning different than that of traditional artificial neural networks (ANNs): SNNs encode data in a series of discrete time spikes, modeled after the action potentials created by neurons in biological brains. The original development of ANNs was also inspired by the human brain, which consists of nearly 100 billion neurons, each of which can have up to 15,000 synapses. Despite this potentially massive parameter space, the human brain only consumes about 20 W and takes up under 3 pounds.  For comparison, today's top-performing language models can require over 300 kWh of power \cite{energy_usage}. Biological neural networks' impressive energy- and space-efficiency is due in part to the fact that most neurons are usually inactive, meaning that the brain is a highly sparse neural network. SNNs attempt to combine more biologically plausible inputs, learning methods, and architectures in order to match the performance of ANNs while also achieving the energy efficiency and robustness of the human brain. Currently, the development of SNNs is a growing field, but lacks clear benchmarks and tool sets, complete optimization of nonlinear spiking dynamics, and robust mechanisms for encoding targets as binary spike trains \cite{time_enc_spikes}. The main goal of this literature review is to chronicle recent advances in SNN optimization, with particular focused applied to biologically plausible mechanisms and advances in the optimization of non-differentiable spike trains.

From an optimization point of view, SNNs pose a unique and interesting challenge, as they lack interpretable models, straight-forward gradient calculation, and also methods that leverage the inherent advantages of SNNs \cite{neural_info_proc} \cite{undst_sn_cxv_opt}.
We have compiled a meta-analysis of recent SNN advances, providing a quantitative comparison between given implementations when possible. Inherent to this process is the identification of standard datasets to act as benchmarks between different implementations.  Another focus of this work is to propose how to better showcase differences and advancements between recent innovations. For performance comparisons, we have identified the most frequently used benchmark datasets so that we can rank existing SNN architectures in an accessible and understandable format.  In a similar vein of work, we have included which tools and frameworks were used to create each model, as there does not exist a standardized set of tools for SNN development (see \cref{table}). We believe such a characterization of tools available to researchers would be a valuable resource for unifying the field, making SNNs more accessible to new researchers.


\section{Foundational Neurobiology} 
This section lays the groundwork for the biological analog from which SNNs originate, focusing on how neurons in the human brain transmit information.

\subsection{Underlying Neuroanatomy}
\textbf{Neurons.} The neuron is the fundamental atomic unit of the brain. In humans, there here are three primary categories of neurons: sensory, motor, and inter-neurons. Neurons usually consist of a bulky soma (i.e., cell body), branching dendrites (which propagate stimuli received from adjacent cells to the soma), and a long axon that forms synaptic junctions with the dendrites of adjacent neurons. 

\textbf{Synapses.} Synapses are the spaces between axon terminals and dendrites through which a presynpatic neuron sends a biological signal to the postsynaptic neuron. Synaptic activity typically consists of either chemical or electrical signals. In the chemical case, presynaptic neuron activation triggers the release of chemical neurotransmitters that bind to receptors lodged in the cell membrane of the postsynpatic neuron, which then cause the propagation of an electrical signal throughout the postsynaptic neuron. In the electrical case, the pre- and post-synpatic neurons are connected by gap junctions which pass the current from one neuron to the other, with the benefit of quicker transmission times \cite{action_potentials}.

\textbf{Action Potentials.} Action potentials (APs) are the discrete signals (spikes) through which biological neurons communicate. Each AP is a continuous chain of depolarization events: one segment along the axon experiences a rise and fall of membrane potential (voltage across the membrane), and this rise and fall is propagated along the axon.  More specifically, APs typically originate from voltage-gated ion channels within a cell's membrane, such that when the cell's membrane potential increases enough to exceed an intrinsic biological threshold, the channels open, allowing charged ions to enter the cell driven by passive gradients, e.g. both a concentration gradient and an electro-chemical gradient due to difference in charges and ion concentrations inside and outside the cell.  Note that the influx of charged ions is equivalent to a current flowing into the cell, and this rapid increase in membrane potential is commonly referred to as depolarization. The resulting diffusion of ions across the cell membrane creates a positive feedback loop for the electro-chemical gradient: incoming ions raise the cell's membrane potential at different points along the membrane, prompting downstream channels to open and allow more ions to enter. Once enough ions have entered the cell, the membrane potential reverses, which prompts the opening of different set of voltage-gated ion channels, this time resulting in an outward flow of ions, repeating the same process as before and bringing the cell membrane back to its resting potential.  The drastic drop in membrane potential is known as hyper-polarization \cite{action_potentials}, \cite{cardiac_action_potentials}.

\begin{figure}[t]
\vskip 0.2in
\begin{center}
\centerline{\includegraphics[width=\columnwidth]{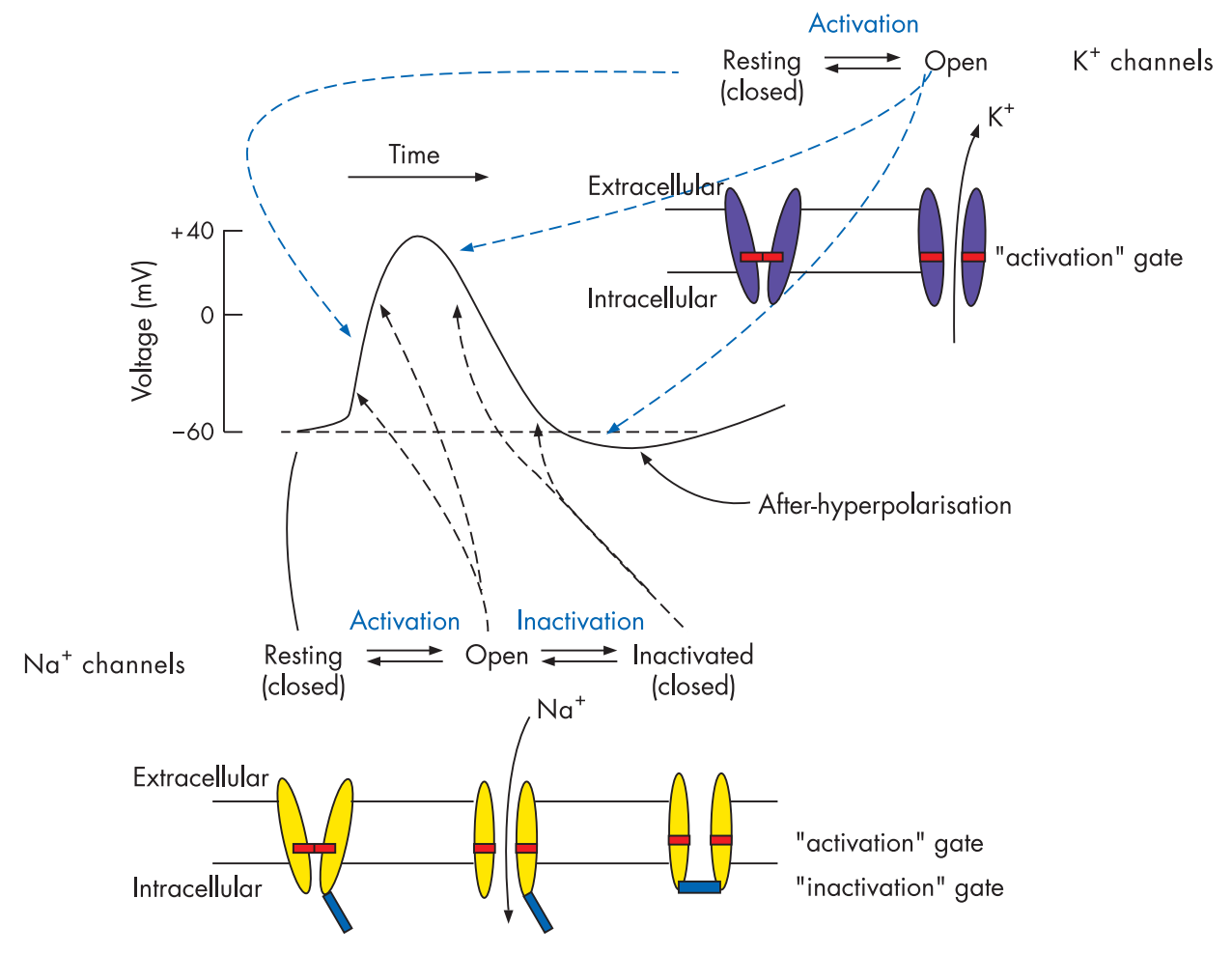}}
\caption{An illustration by \citet{action_potentials} showing the different phases of depolarization and polarization as they relate to action potentials and voltage-gated ion channels.}
\label{fig:Labeled_AP}
\end{center}
\vskip -0.2in
\end{figure}

During action potentials, the membrane potential undergoes very sharp exponential rise and fall within only a few milliseconds. The action potential always drops the neuron's resting potential below the normal resting state potential, resulting in a refractory period (typically about 2 milliseconds) in which the neuron cannot undergo another AP. As will be discussed shortly, these biological mechanisms are the motivation and mathematical framework for today's neural networks. Important terminology borrowed from biology includes the idea of a spike train (the time indices at which spikes occurred for a given neuron) and firing (the creation/propagation of an AP). Note that APs are synonymous with spikes for this discussion \cite{action_potentials}.

\begin{figure}[h!]
\vskip 0.2in
\begin{center}
\centerline{\includegraphics[width=\columnwidth]{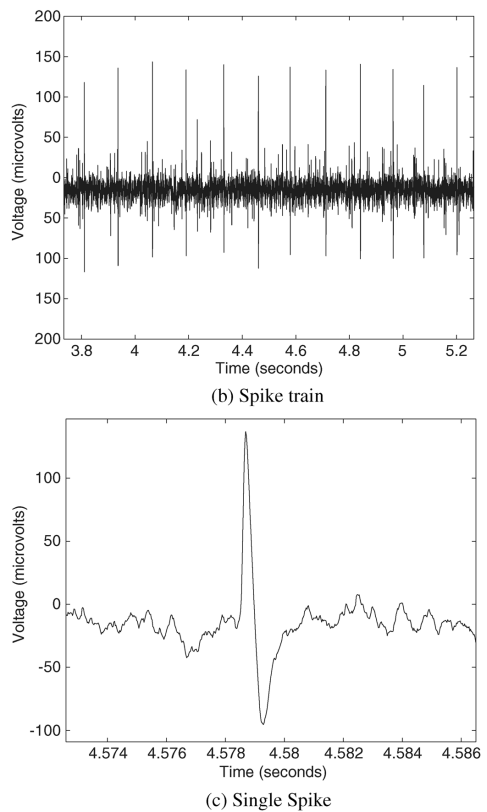}}
\caption{Plotted measurements of (b) an observed spike train and (c) an isolated action potential by \citet{ap_2part_figure}.}
\label{fig:2aps}
\end{center}
\vskip -0.2in
\end{figure}

\subsection{Dynamical Models of the Brain}
\textbf{Hodgkin-Huxley Model.} One of the most popular models for describing the initiation and propagation of action potentials in neurons is the Hodgkin-Huxley (HH) model, as shared by \citet{HH_model}. 
The HH model is a conductance-based dynamical system (see \cref{fig:2aps}), in which each neuron's membrane is modeled simply as a series of parallel capacitive and resistive electrical components, as shown in \cref{eq:hh}. Such abstractions of the intricate inner workings of neurons allow for compartmentalized modeling of many neurons. 

\begin{figure}[h]
\vskip 0.2in
\begin{center}
\centerline{\includegraphics[width=\columnwidth]{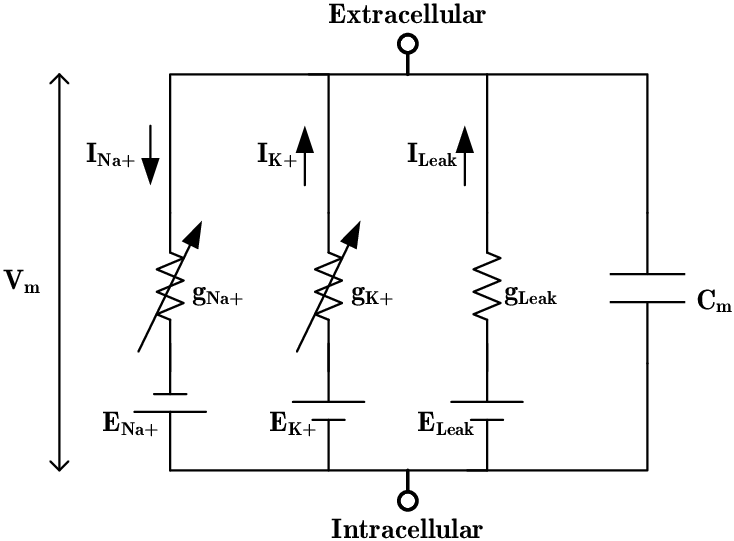}}
\caption{The equivalent circuit of a section of a neuron's membrane, as described by the Hodgkin-Huxley model \cite{hh_figure}.}
\label{fig:2aps}
\end{center}
\vskip -0.2in
\end{figure}

\begin{equation} \label{eq:hh}
    I = C_m \frac{dV_m}{dt} + g_K(V_m - V_K) + g_{Na}(V_m - V_{Na}) + g_l(V_m - V_l)
\end{equation}

The HH model treats each neuron's total current flow $I$ as the sum of four separate current sources described as functions of the membrane potential $V_m$. The first source $C_m \frac{dV_m}{dt}$ arises from the cell membrane separating the neuron's internal contents (negatively charged) from the external environment (positively charged), resulting in the membrane acting as a capacitor with capacitance $C_m$. The other three sources take the form $g_i (V_m - V_i)$, where $g_i$ is the electrical conductance of ion channel $i$ and $V_i$ is the membrane potential threshold at which the direction of current flow in ion channel $i$ reverses (e.g. the electrical gradient is now acting in the opposite direction due to the resulting build up of charge). The specific terms are $i \in [K, Na, l]$, where $K$ and $Na$ represent potassium and sodium ions, and $l$ represents membrane potential leakage instead of a specific ion channel.  Note that other types of currents may also be included, but these are by far the most common.  Recall from the earlier section on neuroanatomy, the first set of opening channels corresponding to the rapid depolarization were the sodium channels, and the second set of channels responsible for bring the cell back to baseline were the potassium channels \cite{HH_model}, \cite{action_potentials}.

Through simple models of neural electrophysiology, scientists can rigorously model the cell membrane, track how it approaches its firing threshold, and therefore understand how best to model the creation and propagation of spikes across neurons. Such understandings yield crucial insights upon which the fundamental ideas of information processing in neural networks can be built. As will be shown, the key similarities between artificial and biological spiking neurons are membrane leakage and communication via action potentials (e.g. spiking activity).

\section{Biological Plausibility and Interpretability} 
This section aims to show various insights into how spiking neural network (SNN) models can be not only human-interpretable, but also biologically plausible \textit{in silico} analogues to biological neural networks.

\subsection{Biological Plausibility}

Traditional artificial neural networks (ANNs) use floating point numbers to encode the information transmission between neurons, whereas biological neurons communicate via action potentials. Given that action potentials occur and decay quickly, they can be approximated as discrete, binary spikes; when taking sparsity into account, information may be transmitted via highly sparse binary vectors instead of relying on arbitrarily precise floating point numbers. However, exactly how neural spike trains encode information is still debated. 
The two most popular information encoding schemes are discussed below.

\textbf{Rate coding.} One of the most popular spike-train information-encoding models is the rate-coding framework, which represents stimuli intensity as a firing rate. If the input stimuli consists of an image, each pixel's intensity could be converted to the firing rate of a Poisson spike train. Rate coding has been experimentally discovered in the visual cortex as well as the motor cortex \cite{NC_in_SNNs}; however, rate coding is not the only method through which neural systems encode information \cite{olshausen2006other}, since it requires long processing periods and therefore has a poor rate of information transmission.

\textbf{Latency coding (TTFS).} An increasingly popular method of information-encoding is the time-to-first-spike (TTFS) model, which only requires a single spike from each neuron in a group to fully encode stimuli. Thus, information can be encoded in a binary fashion over time.  Specifically, TTFS-encoding negatively correlates stimulus intensity with the time that a neuron takes to spike once, and hence is also known as latency-coding. 
The biological plausibility of TTFS-like neural coding has been verified by the dependence on the first spike in systems such as the retina, much of the auditory system, and many responses to tactile stimulus that require fast reactions \cite{NC_in_SNNs}.

\begin{figure}[ht]
\vskip 0.2in
\begin{center}
\centerline{\includegraphics[width=\columnwidth]{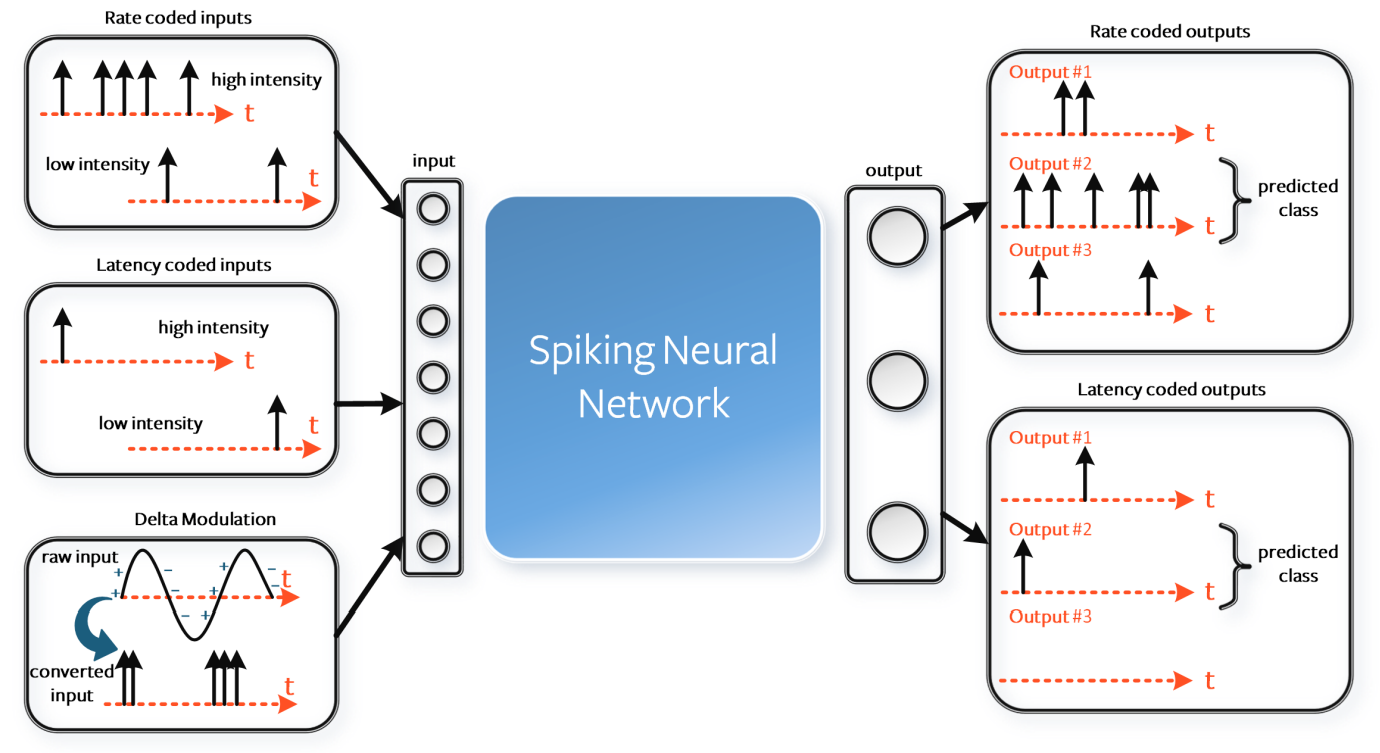}}
\caption{An illustration by \citet{snntorch} showing three common methods for encoding real-valued data as spike trains. Unlike rate- and latency-coding, delta modulation is primarily used only at the sensor-level to represent time-varying stimuli.}
\label{fig:snntorch_coding}
\end{center}
\vskip -0.2in
\end{figure}

\textbf{Neural coding comparisons.} While both rate- and latency-encoding are believed to be used in biological neural computation \cite{olshausen2006other}, SNN practitioners need only be concerned with the advantages of each method: rate coding has a higher error tolerance and provides more information with which gradient learning can converge, while latency coding is significantly more power-efficient and fast than rate coding (since each neuron only spikes once per input) \cite{snntorch}. The choice of optimal coding scheme could depend on whether the dataset is static (e.g., images) or dynamic (e.g., speech or video data). However, multiple works have confirmed that TTFS/latency-coding is the best choice for achieving maximum accuracy while still keeping a low energy consumption \cite{NC_in_SNNs} \cite{snntorch} \cite{HIRE_SNN} \cite{sparse_spike_gd}. \cref{fig:snntorch_coding} depicts the encoding mechanisms of both rate- and latency-encoding. 


\textbf{Leaky Integrate-and-Fire Modeling.} Spiking neural networks are usually modeled and understood as networks of Leaky Integrate-and-Fire (LIF) neurons whose membrane potentials $\mathbf{V}$ evolve at timestep $t$ according to some variation of the following dynamics:
\begin{equation}
    \mathbf{V}[t] = \lambda \mathbf{V}[t-1] + \mathbf{WX}[t] - \mathbf{S}[t]V_{th},
\end{equation}
where $\lambda \in [0, 1)$ is a membrane leakage constant, $\mathbf{X}$ is an input (i.e., an external stimuli to the network or the spiking activity from another neuron), $\mathbf{W}$ is a weight matrix, and the binary spiking function $\mathbf{S}[t] = \begin{cases} 1, & \text{if $\mathbf{V}[t] > V_{th}$} \\ 0, & \text{otherwise} \end{cases}$ is a function of the activation threshold voltage $V_{th}$ \cite{snntorch}.

\subsection{Spiking Neural Networks as Convex Optimizers}

\citet{undst_sn_cxv_opt} tackle SNN biological plausibility from first principles. After showing that SNNs solve quadratic optimization problems, the authors tested their SNNs on a multi-dimensional classification problem, and found that SNNs had irregular spike patterns, a balance of excitation and inhibition, robustness to (Gaussian noise) perturbations, and low firing rates. Furthermore, the authors found that SNNs were robust to cell death (since each neuron corresponds to only one piece of the feasible set boundary), have population firing rates that negatively correlate with the number of neurons.

The authors begin by representing a generic SNN with $N$ neurons and $K$ inputs as follows:
\begin{equation} \label{eq:mancoo_1}
    \dot{\mathbf{V}}(t) = -\lambda \mathbf{V}(t) + \mathbf{Fc}(t) + \mathbf{\Omega s}(t) + \mathbf{I}_{bg}(t),
\end{equation}
where $\mathbf{V}(t) \in \mathbb{R}^N$ represents the voltage of each neuron, $\lambda$ is the time-constant of membrane leakage, $\mathbf{F} \in \mathbb{R}^{N \times K}$ is a feedforward matrix projecting the inputs $\mathbf{c}(t) \in \mathbf{R}^K$ onto the $\mathbb{R}^N$ voltage space, $\mathbf{\Omega} \in \mathbf{R}^{N \times N}$ represents the weights of recurrent connections, $\mathbf{s}(t) \in \mathbb{R}^N$ is a vector containing the neural spike trains, and $\mathbf{I}_{bg}(t) \in \mathbb{R}^N$ represents background currents or noise.

\begin{figure}[ht]
\vskip 0.2in
\begin{center}
\centerline{\includegraphics[width=\columnwidth*3/5]{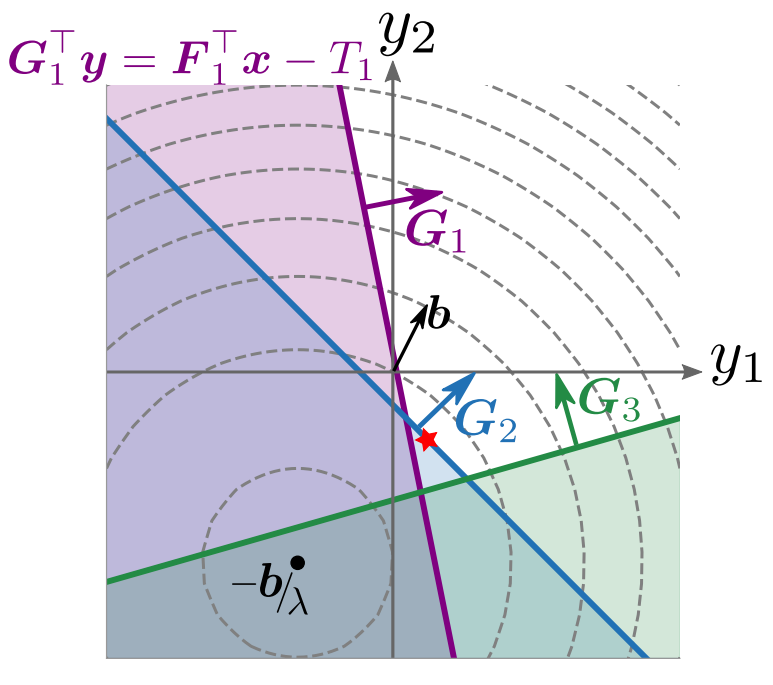}}
\caption{An illustration by \citet{undst_sn_cxv_opt} of a quadratic optimization problem in $\mathbb{R}^2$ with an input $x \in \mathbb{R}^3$. Note how each dimension of the input $\mathbf{x}$ divides the solution space $\mathbb{R}^2$ into two half-planes, one feasible and one infeasible.}
\label{fig:mancoo_2020_1a}
\end{center}
\vskip -0.2in
\end{figure}

Meanwhile, the following is an example of a quadratic optimization problem (or linear if $\lambda = 0$) \cite{convex_optimization}:
\begin{equation} \label{eq:mancoo_2}
    \min_{\substack{\mathbf{y} \\ \mathbf{Fx - Gy} \leq \mathbf{T}}}{\left( E(\mathbf{y}) = \frac{\lambda}{2} \mathbf{y}^T\mathbf{y} + \mathbf{b}^T\mathbf{y} \right)},
\end{equation}
where $\mathbf{y, b} \in \mathbb{R}^M$ are the optimization variable and bias, $\mathbf{x} \in \mathbb{R}^K$ is an input to the problem, and the remaining variables $\mathbf{F} \in \mathbb{R}^{N \times K}$, $\mathbf{G} \in \mathbb{R}^{N \times M}$, $\mathbf{T} \in \mathbb{R}^N$, and $\lambda > 0$ are the parameters of the optimization problem. Note that the optimization problem can be geometrically interpreted as describing a series of $N$ linear inequalities, $\mathbf{G}_i^T \mathbf{y} \geq \mathbf{F}_i^T \mathbf{x} - T_i$, where the column vectors $\mathbf{G}_i$ and $\mathbf{F}_i$ are the $i$-th rows of $\mathbf{G}$ and $\mathbf{F}$, respectively. Each inequality constraint divides the solution space $\mathbb{R}^M$ into two half-planes, with one side satisfying the inequality (i.e., being feasible) and the other not (i.e., being infeasible). Taking the intersection of the feasible set of each inequality constraint yields the feasible set of the optimization problem, as shown in \cref{fig:mancoo_2020_1a} \cite{undst_sn_cxv_opt}.

One algorithm to solve the optimization problem described by \cref{eq:mancoo_2} is gradient descent:
\begin{equation}
    \mathbf{\dot{y}} = -\frac{\partial E}{\partial \mathbf{y}} = -\lambda \mathbf{y} - \mathbf{b}
\end{equation}
However, in order to ensure that $\mathbf{y}$ does not leave the feasible set $\left\{ \mathbf{y} \mid \mathbf{Fx - Gy} \leq \mathbf{T} \right\}$, a term $\mathbf{Ds}(t)$ can be added to $\mathbf{\dot{y}}$ such that the new term reflects $\mathbf{y}$ back into the feasible set whenever $\mathbf{y}$ touches a boundary of the feasible set. Specifically, $\mathbf{s}(t) \in \mathbb{R}^N$ can be modeled a vector of delta-function bouncing events, while the matrix of bounce directions $\mathbf{D} \in \mathbb{R}^{M \times N}$ can be modeled as always bouncing $\mathbf{y}$ orthogonal to the boundary it touched (i.e., $\mathbf{D} \propto \mathbf{G}$):
\begin{equation} \label{eq:mancoo_4}
    \mathbf{\dot{y}} = -\lambda \mathbf{y} - \mathbf{b} + \mathbf{Ds}(t)
\end{equation}
Connecting the quadratic problem formulation to SNNs proceeds by defining the constraint from \cref{eq:mancoo_2} as $\mathbf{V} = \mathbf{Fx} - \mathbf{Gy} \leq \mathbf{T}$, where $\mathbf{V, T}$ represent the neurons' voltages and voltage thresholds:
\begin{align}
    \mathbf{\dot{V}} &= \mathbf{F \dot{x} - G \dot{y}} \\
    &= \mathbf{F \dot{x}} + \lambda \mathbf{Gy + Gb - GDs}(t) \\
    &= -\lambda \mathbf{V + F}(\lambda \mathbf{x + \dot{x}}) - \mathbf{GDs}(t) + \mathbf{Gb},
\end{align}
where the derivation follows from using the definitions of $\mathbf{\dot{V}}$ and $\mathbf{\dot{y}}$ from \cref{eq:mancoo_1,eq:mancoo_4}. The new formulation of $\mathbf{\dot{V}}$ is related to \cref{eq:mancoo_1} in that the inputs $\mathbf{c}(t) = \lambda \mathbf{x + \dot{x}}$, the recurrent connections $\mathbf{\Omega = -GD}$, and the background currents or noise $\mathbf{I}_{bg}(t) = \mathbf{GD}$. The resulting reformulation shows that SNNs solve quadratic ($\lambda > 0$) or linear ($\lambda = 0$) problems.

\section{Optimization}
This section presents the fundamental theory and recent findings in SNN optimization. Key challenges include performing gradient descent on non-differentiable spiking actions and leveraging the intrinsic characteristics of neuromorphic computation towards more efficient SNN training methods.

\subsection{Background}

\textbf{Backpropagation Through Time.} Two of the most prevalent methods for training SNNs are shadow training and backpropagation through time (BPTT). The former involves training a SNN as an approximation of a pre-trained ANN; although such an approach is beneficial for reaching competitive performance in static tasks such as image classification, shadow training will not be the focus of this review due to its inherent inefficiency (requiring two networks to be trained) and its under-utilization of the temporal dynamics of SNNs \cite{snntorch}. The latter method (BPTT) stems from recurrent neural network (RNN) optimization, and becomes necessary for SNN training due to each neuron's membrane potential acting as a form of memory as it decays over time, unlike the activation of ANN neurons.

Following the notation of \citet{snntorch}, the expression for error as a function of SNN weights, $\frac{\partial L}{\partial \mathbf{W}}$, takes the following form when using BPTT:
\begin{equation} \label{eq:BPTT_1}
    \frac{\partial L}{\partial \mathbf{W}} = \sum_t \frac{\partial L[t]}{\partial \mathbf{W}}
\end{equation}
\begin{equation} \label{eq:BPTT_2}
    \frac{\partial L}{\partial \mathbf{W}} = \sum_t \sum_{s \leq t} \frac{\partial L[t]}{\partial \mathbf{W}[s]} \frac{\partial \mathbf{W}[s]}{\partial \mathbf{W}}
\end{equation}
\begin{equation} \label{eq:BPTT_3}
    \frac{\partial L}{\partial \mathbf{W}} = \sum_t \sum_{s \leq t} \frac{\partial L[t]}{\partial \mathbf{W}[s]}, 
\end{equation}
where $L[t]$ and $\mathbf{W}[s]$ are the loss and weights at their respective timesteps $t$ and $s$. \cref{eq:BPTT_2} follows from \cref{eq:BPTT_1} since $\mathbf{W}[s]$ directly determines $L[s]$ (referred to as an \textit{immediate influence}) while also influencing $L[t]$ for all $t > s$ (referred to as a \textit{prior influence}). \cref{eq:BPTT_2} then simplifies to \cref{eq:BPTT_3} since a recurrent system constrains the weights $\mathbf{W}$ to be shared along all timesteps ($\mathbf{W}[0] = \mathbf{W}[1] = ... = \mathbf{W}$), meaning that $\frac{\partial \mathbf{W}[s]}{\partial \mathbf{W}} = 1$.

\textbf{Surrogate Gradients.} Unlike ANNs, SNNs use spikes to perform computation. Letting $\mathbf{S}$ denote spiking activity, calculation of the gradient $\frac{\partial L}{\partial \mathbf{W}}$ takes the form of $\frac{\partial L}{\partial \mathbf{S}}$ multiplied by additional terms (via the chain rule) that constitute $\frac{\partial \mathbf{S}}{\partial \mathbf{W}}$. However, since the spiking activity $\mathbf{S}$ as a function of synaptic weights $\mathbf{W}$ involves a Heaviside step function, the derivative $\frac{\partial \mathbf{S}}{\partial \mathbf{W}}$ contains an ill-behaved Dirac-delta function. As a relaxation of the non-smooth spiking nonlinearity, surrogate gradients are used as an alternative to the derivative of the Heaviside step function during backpropagation \cite{https://doi.org/10.48550/arxiv.1901.09948}. The choice of surrogate gradient for SNN optimization is not unique, but recent work has shown that SNNs are robust to the choice of surrogate gradient \cite{the_remarkable_robustness}.

\begin{figure}[ht]
\vskip 0.2in
\begin{center}
\centerline{\includegraphics[width=\columnwidth]{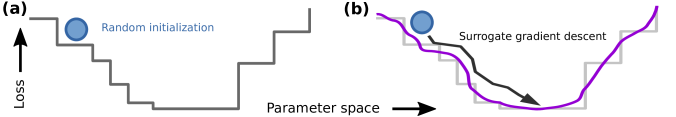}}
\caption{An illustration by \citet{https://doi.org/10.48550/arxiv.1901.09948} showing how surrogate gradients can be used to smoothen the loss landscape of SNNs.}
\label{fig:surrogate_gradients}
\end{center}
\vskip -0.2in
\end{figure}

\subsection{Advancements in Surrogate Gradient Descent}
\textbf{Differentiable Spike.} \citet{neural_info_proc} have recently brought attention to the finite difference gradients (FGD) as a novel method of computing surrogate gradients. With loss $L$ and network parameters $\mathbf{W}$, approximation of the true gradient $\nabla_\mathbf{W} L$ via the FGD $\hat{\nabla}_{\varepsilon, \mathbf{w}} L$ is defined as:
\begin{equation}
    \hat{\nabla}_{\varepsilon, \mathbf{w}} L = \frac{1}{2 \varepsilon} \begin{bmatrix}
        L(\mathbf{w}_1 + \varepsilon \mathbf{e}_1) - L(\mathbf{w}_1 - \varepsilon \mathbf{e}_1) \\
        L(\mathbf{w}_2 + \varepsilon \mathbf{e}_2) - L(\mathbf{w}_2 - \varepsilon \mathbf{e}_2) \\
        ... \\
        L(\mathbf{w}_k + \varepsilon \mathbf{e}_k) - L(\mathbf{w}_k - \varepsilon \mathbf{e}_k)
    \end{bmatrix},
\end{equation}
where $\textbf{w} \in \mathbb{R}^k$ is the flattened vector of $\mathbf{W}$, $\mathbf{e}_i$ is the standard basis vector with 1 at its $i$-th index, and $\varepsilon$ is a hyperparameter. As $\varepsilon \rightarrow 0$, $\hat{\nabla}_{\varepsilon, \mathbf{w}} L$ becomes the true gradient $\nabla_\mathbf{W} L$. \citet{neural_info_proc} have shown that, although using very small $\varepsilon \in [0.1, 0.001]$ successfully approximates the true gradient in artificial neural networks, SNN optimization is more sensitive to the magnitude of $\varepsilon$, since the optimal values of $\varepsilon$ must be low enough to faithfully approximate the characteristics of the SNN gradient while being high enough so as to have a smoothening effect on the discontinuous SNN loss landscape.

\citet{neural_info_proc} then proceed to define the Differential Spike (Dspike) surrogate gradient with temperature parameter $b$ and input $x$:
\begin{equation}
    \text{Dspike}(x, b) = \begin{cases}
        1, & \text{if $x > 1$} \\
        \frac{\text{tanh}(b(x - 0.5)) + \text{tanh}(\frac{b}{2})}{2 \cdot \text{tanh}(\frac{b}{2})} & \text{if $0 \leq x \leq 1$} \\
        0, & \text{if $x < 0$}
    \end{cases}
\end{equation}
The authors train each of their SNNs by, in each epoch $e$, first computing the FDG $\hat{\nabla}_{\varepsilon, \mathbf{w}} L$ of the first SNN layer, then choosing the value of $b \in \{ b^{e-1}, b^{e-1} + \Delta b, b^{e-1} - \Delta b \}$ (where $\Delta b$ is a step size hyperparameter and $b^{e-1}$ is the optimal $b$ from the previous epoch) to maximize the cosine similarity between $\hat{\nabla}_{\varepsilon, \mathbf{w}} L$ and $\nabla_{b, \mathbf{w}} L$, and finally optimizing the entire SNN using Dspike gradient descent with the newly optimized value of $b$.

\textbf{Information-Maximization and ESG.} More recently, \citet{IMLoss} have introduced Information-Maximization Loss (IM-Loss) and the Evolutionary Surrogate Gradient (ESG) for differentiable spike activity estimation. The authors derive the IM-Loss by first arguing that one key reason why SNN performance lags behind ANN performance is because of the loss of information produced by quantizing membrane voltages into binary spiking functions. Representing the tensor of membrane potentials as $\mathbf{V}$ and the tensor of spiking activity as $\mathbf{S}$, the authors seek to maximize the mutual information between $\mathbf{v}$ and $\mathbf{S}$ by maximizing the entropy of $\mathbf{S}$:
\begin{equation}
    \argmax_{\mathbf{V}, \mathbf{S}} I(\mathbf{V}; \mathbf{S}) = H(\mathbf{S}) - H(\mathbf{S} | \mathbf{V}) = H(\mathbf{S}),
\end{equation}
where $H(\mathbf{S} | \mathbf{V})$ is argued to be 0 due to the binary nature of $\mathbf{S}$. Since each element of $\mathbf{S}$ is either 0 or 1, the entropy of $\mathbf{S}$ is maximized when each element of $\mathbf{S}$ is equally likely to be 0 or 1. Using the observation that $\mathbf{V}$ is often Gaussian distributed and therefore has a median equal to its mean $\mathbf{\bar{V}}$, \citet{IMLoss} propose the IM-Loss as: 
\begin{equation}
    \mathcal{L}_{IM} = \frac{1}{L} \sum_{l=0}^L (\bar{V}^{(l)} - V_{th})^2,
\end{equation}
where $V_{th}$ is the spike activation threshold voltage. The authors' final loss function for SNN training was $\mathcal{L}_{Total} = \mathcal{L}_{CE} + \lambda \mathcal{L}_{IM}$, where $\mathcal{L}_{CE}$ is cross-entropy loss and $\lambda = 2$.

Similarly to Dspike \cite{DSpike}, \citet{IMLoss} use a differentiable spike activation function $\varphi(x)$ whose sole parameter $K(i)$ evolves as a function of the epoch index $i \in [0, N - 1]$:
\begin{align}
    \varphi(x) &= \frac{1}{2} \text{tanh}(K(i) (x - V_{th})) + \frac{1}{2} \\
    K(i) &= \frac{1}{9} ( (10^\frac{i}{N} - 1) K_{max} + (10 - 10^\frac{i}{N}) K_{min} )
\end{align}
Since $K(i)$ grows monotonically as a function of the epoch index $i$, the Evolutionary Surrogate Gradient $\varphi(x)$ aims to improve SNN convergence by beginning as a function with a very wide gradient (i.e., allowing many weights to be updated) and then gradually transitioning to a function that more closely resembles the true spiking function.

\subsection{Leveraging Sparsity}
\citet{sparse_spike_gd} present a novel approximation of surrogate gradient descent that leverages spike train sparsity for remarkable improvements in computational efficiency and speed. The authors begin with the assumption that neurons operate according to a simplified Leaky Integrate and Fire model:
\begin{equation}
    V_i^{(l)}[t] =
        \begin{cases}
            F(V_i^{(l)}[t-1], x[t]), & \text{if $V_i^{(l)}[t] < V_{th}$} \\
            V_r, & \text{if $V_i^{(l)}[t] \geq V_{th}$}
        \end{cases}
\end{equation}
$F$ represents the dynamics governing the voltage $V_i^{(l)}$ of the $i$-th neuron of layer $l$, $V_{th}$ is a voltage threshold, and $V_r$ is the resting potential of the neuron after spiking. At each time, neurons spiking according to $S_i^{(l)}[t] = f(V_i^{(l)}[t])$, where $f(v) \in \{0, 1\}$ equals $1$ if and only if $v > V_{th}$. The authors use gradient descent via backpropagation through time (BPTT) with the SuperSpike surrogate gradient $g(V) = \frac{1}{ (\beta \lvert V - V_{th} \rvert + 1)^2}$ \cite{superspike, the_remarkable_robustness}. In order to make gradient optimization more efficient, the authors introduce an approximation of the surrogate gradient:
\begin{equation}
    \frac{dS_i^{(l)}[t]}{dV_i^{(l)}[t]} =
        \begin{cases}
            g(V_i^{(l)}[t]), & \text{if neuron $i^{(l)}$ is active} \\
            0, & \text{otherwise}
        \end{cases}
\end{equation}
where the $i$-th neuron of layer $l$ is defined as active if and only if
\begin{equation}
    \lvert V_i^{(l)} - V_{th} \rvert < B_{th}
\end{equation}

By approximating the surrogate gradient as being nonzero only for neurons that are active at any timestep $t$, the resulting derivation of BPTT yields an optimization strategy that approximates weight updates very well, having a backward pass speedup of up to 150x and saving up to 85\% more memory compared to vanilla BPTT on datasets such as Fashion-MNIST \cite{fashion-mnist} (processed via latent coding), Neuromorphic-MNIST \cite{n-mnist}, and Spiking Heidelberg Digits \cite{Cramer_2022} \cite{sparse_spike_gd}.

\subsection{Learning Rules}
One of the most common learning rules for training SNNs is backpropagation modeled as the reverse of a simple forward pass. However, backpropagation methods, such as backpropagation through time (BPTT), are not biologically plausible \cite{Hunsberger_Eric2018}, are memory-intensive, and are not fully compatible with neuromorphic hardware \cite{event-driven_backprop}, instead necessitating computationally intensive training on normal hardware. This section explores recent extensions or alternatives to backpropagation.

\textbf{Spike-Timing Dependent Plasticity.} Another alternative to BPTT that focuses on spike timing is spike-timing dependent plasticity (STDP), a learning framework whose origins are not in optimization, but rather in experimental observations of neural plasticity. Unlike event-driven optimization, STDP is unsupervised and does not use gradient descent or backpropagation, instead determining the synaptic connection strength between each neuron solely as a function of the relative timings between pre- and post-synaptic neuronal spikes \cite{snntorch}. Such learning rules are commonly known as Hebbian learning methods \cite{caporale2008spike} and are characterized by the principle that neurons that fire together wire together \cite{snntorch}. 
\begin{equation} \label{eq:stdp}
    \Delta w_{ij} =
    \begin{cases}
        a^+ w_{ij} (1 - w_{ij}), & \text{if $t_j - t_i \leq 0$} \\
        a^- w_{ij} (1 - w_{ij}), & \text{if $t_j - t_i > 0$}
    \end{cases}
\end{equation}
One example of STDP learning rules for SNNs is shown in \cref{eq:stdp}, where $a^+$ and $a^-$ are learning rates, $t_i$ and $t_j$ are the post- and pre-synaptic spike times of two neurons, and $w_{ij}$ is the strength of connection between the two neurons \cite{Kheradpisheh}. A recent extension of this STDP is R-STDP, or reward-modulated STDP, where reward/punishment signals are used in conjunction with traditional STDP to train the weights \cite{STDP}.

\textbf{Event-Driven Optimization.} One potential alternative to traditional backpropagation through time for SNNs is event-driven optimization, specifically backpropagation with respect to spike timing \cite{spikeprop}. Such optimization methods involve calculating time-based gradients that indicate whether a spike should happen earlier or later with respect to time. \citet{event-driven_backprop} have recently been the first to successfully train SNNs on the CIFAR-100 dataset using event-driven methods.

\textbf{Online Training Through Time.} \citet{online_training_through_time} recently proposed another alternative to traditional backpropagation through time. Their method, referred to as online training through time (OTTT), diverges from traditional SNN optimization by not using a surrogate gradient. Instead, the authors leverage the fact that the gradient of spiking activity is naturally zero almost everywhere to approximate $\frac{\partial \mathbf{V}^{(l)}[t+1]}{\partial \mathbf{S}^{(l)}[t+1]} \frac{\partial \mathbf{S}^{(l)}[t+1]}{\partial \mathbf{V}^{(l)}[t]} \approx 0$, where $\mathbf{V}^{(l)}[t]$ and $\mathbf{S}^{(l)}[t]$ are the voltage and spiking activity of layer $l$ at time $t$. As a result, each layer's voltage evolves simply as a function of its membrane leakage constant $\lambda < 1$: $\frac{\partial \mathbf{V}^{(l)}[t+1]}{\partial \mathbf{V}^{(l)}[t]} = \lambda \mathbf{I}$. As a result, \citet{online_training_through_time} are able to use a loss function and a gradient whose values are instantaneous (i.e., only depending on the timestep $t$). Further analysis formulates OTTT as a three-factor Hebbian learning rule, thus building a pathway towards biologically plausible training on neuromorphic devices.

\textbf{Implicit Differentiation.} As another alternative to feedforward SNNs trained via BPTT, \citet{implicit_differentiation} bring to attention feedback spiking neural networks (FSNNs), which are SNNs with recurrent connections, and propose implicit differentiation on the equilibrium state (IDE) as a method to train them. The authors define the equilibrium state $\mathbf{a}^*$ of a network $f_\theta$ with parameters $\theta$ as the state where $\mathbf{a}^* = f_\theta (\mathbf{a}^*)$. Implicit differentiation on the equation satisfies $(\mathbf{I} - \frac{\partial f_\theta ( \mathbf{a}^* )}{\partial \mathbf{a}^*}) \frac{d \mathbf{a}^*}{d \theta} = \frac{\partial f_\theta( \mathbf{a}^* )}{\partial \theta}$. Writing the objective function evaluated at $\mathbf{a}^*$ as $\mathcal{L}(\mathbf{a}^*)$, and letting $g_\theta (\mathbf{a}) = f_\theta (\mathbf{a}) - \mathbf{a}$, differentiation of $\mathcal{L}(\mathbf{a}^*)$ with respect to the parameters $\theta$ can be expressed as:
\begin{equation}
    \frac{\partial{\mathcal{L}(\mathbf{a}^*)}}{\partial \theta} = -\frac{\partial{\mathcal{L}(\mathbf{a}^*)}}{\partial \mathbf{a}^*} \left( \mathbf{J}^{-1}_{g_\theta} |_{\mathbf{a}^*} \right) \frac{\partial f_\theta(\mathbf{a}^*)}{\partial \theta},
\end{equation}
where $\left( \mathbf{J}^{-1}_{g_\theta} |_{\mathbf{a}^*} \right)$ is the inverse Jacobian of $g_\theta$ evaluated at $\mathbf{a}^*$. Solving the inverse Jacobian requires solving an alternative linear system $\left( \mathbf{J}^T_{g_\theta} |_{\mathbf{a}^*} \right) \mathbf{x} + \left( \frac{\partial \mathcal{L}(\mathbf{a}^*)}{\partial \mathbf{a}^*} \right)^T = 0$, which can be done using second-order quasi-Newton methods or by using a fixed-point update scheme $\mathbf{x} = \left( \mathbf{J}^T_{f_\theta} |_{\mathbf{a}^*} \right) \mathbf{x} + \left( \frac{\partial \mathcal{L}(\mathbf{a}^*)}{\partial \mathbf{a}^*} \right)^T$, since $\left( \mathbf{J}^T_{g_\theta} |_{\mathbf{a}^*} \right) = \left( \mathbf{J}^T_{f_\theta} |_{\mathbf{a}^*} \right) - \mathbf{I}$ \cite{implicit_differentiation}. Updating the parameters $\theta$ can then occur via gradient descent based on $\frac{\partial \mathcal{L}}{\partial \theta}$.

\section{Energy Efficiency}
This section focuses on how SNNs garner impressive energy savings, covering the mathematical underpinnings as well as documenting notable gains in memory and complexity reduction.

\subsection{Low-Power Applications and Constraints}
SNNs hold particular promise for low-power and implantable applications, such as brain-machine interfaces (BMIs), where there exists a strict power requirement: chips directly in contact with brain tissue cannot emit more than 10 mW of energy before permanently damaging brain cells by heating them by over 1 degree Celsius \cite{Shenoy}.  SNNs promise to reduce energy consumption and thus heat output, making them a promising choice for implants \cite{kid_named_finger}. However, one can only benefit from SNNs when running on hardware built to take advantage of the inherent benefits of SNNs. such as their sparsity and inclination for parallel computation. Neuromorphic processors deviate from the traditional von Neumann architecture of today's general purpose CPUs by employing massive parallelism, asynchronous event-driven operations, and a more highly-distributed, easily-accessible memory that reduces the number of computations---and thus, the energy consumption---required for information processing \cite{NC_in_SNNs}. As found by \citet{Shenoy}, neuromoprhic hardware implementations can result in SNNs which consume as little as 0.1 mW, well below the aforementioned maximum power threshold of 10 mW for BMIs.

\subsection{Mechanisms for Energy Savings}
\textbf{Reduced Activity.} Energy consumption has been shown to be reduced by up to 3 orders of magnitude when compared to conventional ANN implementations, largely due to spiking activity, sparsity, and static data suppression (event-driven processing) \cite{186} \cite{188}. Neurons producing discrete spikes in sparsely-connected networks result in spatio-temporally sparse activity, where activity can be defined as the following:
\begin{equation}
    A = \frac{100a}{BTN},
\end{equation}
where $a$ is the number of active neurons, $B$ is the batch size, $T$ is the total number of time steps, and $N$ is the number of neurons in the layer \cite{sparse_spike_gd}. As one would intuitively guess, the level of activity is inversely proportional to the sparsity of the inputs: as more neurons become active such that the fraction of neurons firing, $a/N$, increases, the network is by definition less sparse.
The energy-efficiency of spike-based computation becomes especially apparent when considering that only a fraction of neurons in real brains use rate-coding, whereas all others opt for much sparser representations of activity \cite{olshausen2006other}. 

\textbf{Estimating Computational Expenditure.} Computational energy for neural networks is typically assessed in two ways: the number and type of operations performed, and the frequency of memory access. For the former, many papers have quantified the energy cost of AC (accumulation) and MAC (multiply and accumulate) operations for various CMOS processors.  In ANNs, the summation of the weighted inputs to a neuron require a MAC operation (e.g. one MAC operation per input), whereas for SNNs, spikes rely on the much cheaper AC operation per input. For comparison, one conservative estimate is that three AC operations equate to one MAC operation \cite{kid_named_finger}. Note that SNNs need to update their membrane potentials every time step, accounting for an additional MAC operation per each neuron, as opposed to ANNs where there is a MAC operation per input to each neuron.

\textbf{Memory Bottleneck.} As for memory, the continual increase of neural network parameter complexity produces an increasing difficulty of storing everything on-chip: when accessing memory that is further away (for example, there is a substantial jump in delay for accessing larger memory bases; instead of requiring data from the CPU's cache, having to read from RAM, or the SSD), the time and energy required grows, thus presenting a key bottleneck in the implementation of all neural network architectures \cite{Dataflows}. Many popular learning methods in today's SNNs (e.g., BPTT variants) are biologically implausible, since they require access to past states to use as inputs (i.e., data must be pulled from memory): thus, the memory consumed scales with time, limiting BPTT to small temporal dependencies \cite{snntorch}. Another example of this hardware bottleneck is that, in dense back propagation situations, there can only be a few thousand time steps, as gradients must be computed at each of those time steps. Notably, many of these gradients have little effect on the output; \citet{sparse_spike_gd} present a backpropagation technique that only computes the gradients of active neurons (i.e., only calculates the gradients that meaningfully affect the output), which facilitates training on GPUs, whereas traditional BPTT runs out of memory before it can converge. As \citet{sparse_spike_gd} point out, this effectively means that such SNN implementations can be trained with more data for a longer amount of time when compared to previous generations of SNNs, given the same computing resources. 

\subsection{Observed Energy Efficiency Gains} 
An observed issue with reporting the energy efficiency or decrease in memory/energy consumption by SNNs throughout the literature was that there were no standardized measurements for quantifying the energy savings offered by SNNs. A number of works reported the number of estimated MAC and AC/add operations that their implementation required, frequently contrasted against the number of said operations required for ANN equivalents; while this method is simpler to estimate on the researcher's behalf, it remains unstandardized, as different researchers estimate the number of operations in different ways, and the number of operations is fundamentally a proxy for true energy consumption. As for the works that do report the expected energy consumption, different teams make different assumptions about how many AC operations are equal to one MAC operation, and thus introduce a wide range of reported energy savings. We recommend that researchers take the conservative approach, using the methodology suggested by \citet{3MACs} of using 3 additions as the equivalent to 1 MAC operation.  Note that while this is of course a proxy for actual energy consumption, the amount of energy consumed is also dependent on the actual hardware used to train and deploy the model.

\textbf{HIRE-SNN.}
HIRE-SNN VGG11 (a direct-coded SNN) compared against a rate-coded SNN and an ANN achieved 4.6x and 10x less energy consumption, respectively \cite{HIRE_SNN}.

\textbf{Sparse Spiking Gradient Descent.}
\citet{sparse_spike_gd} introduced a method that speeds up backward execution by up to two orders of magnitude, and thus they expected that this method would likewise reduce the energy consumption by a similar factor. By constraining back propagation to active neurons, 98\% of the neurons (i.e., 98\% of the gradient calculations) can be skipped. Such accelerations yield much faster back propagation times, lower memory requirements, and less energy consumption on GPU hardware while not affecting test accuracy. These laudable improvements are possible because the surrogate gradient used to approximate the derivative of the spiking function is larger when the membrane potential is closer to the threshold, resulting in most of the gradient being concentrated on active neurons \cite{sparse_spike_gd}.

\textbf{SNN Brain-Machine Interface.}
\citet{kid_named_finger} compared ANNs and SNNs for an implantable BMI that decodes finger velocity. They found that an ANN required 529K total operations and 2116K memory accesses, and an SNN trained to approximate the ANN similarly required 293K total operations and 2615K memory accesses. However, the authors' proposed SNN required just 36K total operations and only 199K memory accesses. Their proposed SNN produced reductions of ~95\% and ~90\% in the number of total operations required, and ~91\% and ~93\% in the number of memory accesses, respectively \cite{kid_named_finger}.

\section{Best Practices}
This section aims to present the best practices in SNN training and evaluation: widely used benchmarks, baseline standards, architectures, and frameworks.

\subsection{Benchmarks/Datasets}

\textbf{Background.} Throughout the development of ANNs, certain datasets---such as the MNIST \cite{mnist} and CIFAR datasets \cite{cifar}---have served as reliable benchmarks for all researchers to compare against. The development of SNNs has come as a subset of the growing neuromorphic engineering landscape, wherein advances made in software and hardware must be accompanied by advances in benchmarks such as standardized neuromorphic datasets.   
A key prerequisite to SNN implementation is to encode the input data as spikes: thus, event-based sensors (such as the ATIS vision sensors) that record pixel-level spike-equivalents of the data in the form of inherently neuromorphic datasets have gained traction, as have software conversions of traditional datasets into neuromorphic form. 
Note that, as discussed earlier, the precise mechanism through which to encode data as binary spike trains is not unique, with the primary two modes of doing so being rate-coding and latency-coding.

\textbf{Neuromorphic Datasets.} Many imaging datasets now have neuromorphic counterparts, wherein the changes in intensity for each pixel (measured from 3 vantage points via sensors) are represented as spikes occurring at the given pixel locations \cite{neuromorphic_datasets}. The most common neuromorphic datasets in the literature include the N-MNIST dataset (spiking version of the classic MNIST dataset), DVS Gestures (video data of hand gestures recorded under different lighting conditions), and SHD (spiking version of the Heidelberg Digits in the form of audio data converted via a simulated cochlear model). A number of other neuromorphic vision/imaging datasets exist, such as ASL-DVS, DAVIS, MVSEC, POKER DVS, and DSEC, and other common audio datasets simulating cochlear activity include SSC and N-TIDIGITS \cite{snntorch}. 

\textbf{Non-Neuromorphic Datasets.} Despite the proliferation of neuromorphic datasets, they remain niche and unstandardized, indicating that many practitioners prefer to convert non-neuromorphic datasets to spiking inputs themselves.  That said, there are a number of commonly-used datasets, particularly the MNIST and CIFAR-10 datasets. Of the total benchmarks reviewed, 26.6\% of all models were compared using the MNIST dataset (33.1\% when including MNIST variants such as N-MNIST and F-MNIST), and CIFAR-10 was the second most common with 30.5\% of models being benchmarked against it. Notably, while more papers choose to use MNIST, the papers reviewed that used CIFAR tended to have many more models per paper. It is important to acknowledge that many SNN implementations utilized in neuroengineering applications are trained and deployed directly onto private collected data (for instance, EEG data collected from an implant within a rhesus monkey) as opposed to a standardized benchmark, and thus these were not included in the prior calculations.

\textbf{Recommendations.} We recommend that aspiring researchers use either the MNIST dataset or CIFAR-10, as they remain the most common and thus have the benchmarking network effect. However, in a further effort to standardize results across the field, we also recommend using the publicly available N-MNIST or DVS Gestures datasets, as these would help ensure repeatability since these latter two datasets are already converted to a neuromorphic input form.

\textbf{Baselines.} One primary result of our literature review is a compilation of some of the latest state of the art baselines in SNN optimization, which we have compiled in \cref{table}.

\subsection{Architectures and Frameworks}

\textbf{Software.} There exists a plethora of development options for implementing SNNs, each with subtle differences, but many attempting to fill the same niche. Common frameworks include BindsNET, Norse, Nengo, SpykeTorch, and SNNToolBox.  BindsNET is built on top of PyTorch as is focused on simulating SNNs for general machine learning and reinforcement learning purposes \cite{BindsNET}. Norse similarly expands PyTorch with SNNs \cite{Norse}. Nengo is built on top of Keras/TensorFlow, and has specific modules for deep learning and simulation backends (e.g. on OpenCL, FPGAs, Loihi) \cite{NengoDL}. SpykeTorch is also built from PyTorch, and is focused on simulating convolutional SNNs with one spike per neuron, rank-order information encoding, and learning via STDP or R-STDP \cite{SpykeTorch}. Finally, SNNToolBox (SSN-TB) is a framework solely focused on transforming traditional ANNs to SNNs and supports a number of different encoding schemes: SNN-TB can be used with models from Keras/TF, PyTorch, and other popular deep learning libraries, and provides an interface for both simulation backends (pyNN, brian2, etc.) and deployment (via SpiNNaker or Loihi) \cite{snn-tb}. Many modern machine learning frameworks do not offer ways to accelerate the convolution of binary spike activation, and furthermore struggle to implement custom continuous-time differential expressions given the inherent discrete execution of their built-in operations \cite{DSpike} \cite{IMLoss}.

\textbf{Hardware.} Another equally important aspect of SNN implementation is the hardware itself: when implemented on non-neuromorphic hardware (i.e., traditional von-Neumann architectures), the benefits of SNNs are less pronounced since the hardware is physically incapable of taking advantage of the massive parallelism offered by SNNs. By far, the most common hardware for SNNs is the Loihi chip by Intel (whose successor, the Loihi 2, was recently released in late 2021). The Loihi is a neuromorphic many-core processor that supports on-chip learning, making it a viable option for both inference and deployment. The Loihi has 128 cores, simulates up to 131,072 neurons with 130,000,000 synapses possible; furthermore, each Loihi can be put in parallel with up to 16,384 other chips, reaping the benefits of massive parallelism by allowing the number of effective on-chip cores to be 4,096 \cite{Loihi}. Another common implementation tool for SNNs is SpiNNaker, a million-core, open-access, ARM-based neuromorphic computing platform developed and housed in the University of Machester: SpiNNaker is used for simulation and testing for applications that do not require on-site implementations \cite{spinnaker}.

\textbf{Architectural Techniques.} The final, but just as important, subsection of methodologies required to enable SNNs to reach their potential as universal approximators \cite{theoretically_provable_snns} consists of how practitioners can best structure, initialize, or normalize the connections between spiking layers to maximize performance. As a prominent example, \citet{snn_resnet} have recently developed a spike-element-wise (SEW) residual network that faithfully implements key properties of ANN residual networks \cite{resnet} in SNNs: namely, the ability to learn an identity function and the ability to train very deep SNN architectures (over 100 layers) by virtue of mitigating vanishing/exploding gradient problems. New work \cite{fluctuation-driven_initialization} in SNN initialization has extended weight initialization methods that target a specific variance for neuronal activity---such as Glorot and He initialization \cite{glorot_initialization, he_initialization}---to spiking networks. Specifically, \citet{fluctuation-driven_initialization} present a new SNN initialization formula that surpasses He initialization for deep SNNs (i.e., more than 7 layers) by simply targeting a specific membrane potential variance. As for SNN normalization, one of the more fundamental works is threshold-dependent batch normalization (tdBN) \cite{tdbn}, which extends traditional batch normalization (BN) \cite{batch_norm} to the additional temporal dimension brought by SNNs and makes the normalized pre-activation variance dependent on the firing threshold $V_{th}$. Newer developments include batch normalization through time (BNTT) \cite{bntt} and temporal effective batch normalization (TEBN) \cite{tebn}, which both aim to mitigate temporal co-variate shift of spiking activity.

\section{Conclusion}
Our review has covered the fundamental principles and the cutting-edge methods for reaching competitive performance in spiking neural networks (SNNs) while also improving SNN memory and energy efficiency.
This literature search has identified the most common benchmarks for aspiring SNN practitioners to use, the current state of the art performances on those benchmarks, and the trajectories of recent advances in SNN optimization. SNNs offer an exciting new technology, particularly for low-power, implanted, mobile, or other hardware-constrained applications, such as brain-machine interfaces and wearable technologies.  Our understanding of SNNs as the third generation of neural networks continues to grow; with such growth comes not only the potential to improve neural network performance while reducing energy costs, but also the potential to further develop our understanding of neuroscience and neural computation.

\section*{Acknowledgements}

We thank Anastasios Kyrillidis for his guidance and support during the writing process.

\bibliography{_main}
\bibliographystyle{icml2022}


\comment{
    \newpage
    \appendix
    \onecolumn
    \section{You \emph{can} have an appendix here.}
    
    You can have as much text here as you want. The main body must be at most $8$ pages long.
    For the final version, one more page can be added.
    If you want, you can use an appendix like this one, even using the one-column format.
}

\newpage
\appendix
\onecolumn
\section{Performance Comparisons}

\begin{tiny}
\begin{table}[h] \label{table}
    \centering
    \begin{tabular}{|p{1.25in}p{0.9in}p{1.4in}p{0.9in}p{0.8in}p{0.45in}|} 
    \hline
        Model & Architecture & Learning Method & Training Framework & Dataset & Accuracy \\ \hline
        \textbf{\cite{sparse_spike_gd}} & \textbf{3 layer SNN} & \textbf{Backpropagation} & \textbf{Pytorch CUDA Extension} & \textbf{SHD} & \textbf{77.5} \\
        \cite{sparse_spike_gd} & 3 layer SNN & Sparse Backpropagation & Pytorch CUDA Extension & SHD & 71.7 \\ \hline

        \cite{Diehl} & 2 layer SNN & Unsupervised & BRIAN & MNIST & 95.0 \\
        \cite{Liu} & 3 layer SNN & Unsupervised & BRIAN & MNIST & 81.9 \\ 
        \cite{Kulkarni} & 3 layer SNN & Supervised & BRIAN & MNIST & 98.0 \\
        \textbf{\cite{Kheradpisheh}} & \textbf{6 layer SNN} & \textbf{Supervised} & \textbf{BRIAN} & \textbf{MNIST} & \textbf{98.4} \\
        \cite{NC_in_SNNs} & 2 layer SNN & TTFS Encoding, STDP & Not Listed & MNIST & 88.6 \\ 
        \cite{NC_in_SNNs} & 2 layer SNN & Phase Encoding, STDP & Not Listed & MNIST & 88.2 \\ 
        \cite{NC_in_SNNs} & 2 layer SNN & Burst Encoding, STDP & Not Listed & MNIST & 88.4 \\
        \cite{NC_in_SNNs} & 2 layer SNN & Rate Encoding, STDP & Not Listed & MNIST & 87.5 \\ 
        \cite{Rathi} & 2 layer SNN & Unsupervised STDP & BRIAN & MNIST & 93.2 \\
        \cite{Rathi} & 4 layer SNN & Unsupervised STDP & BRIAN & MNIST, TI46 & 98.0 \\ \hline
        
        \cite{sparse_spike_gd} & 3 layer SNN & Backpropagation & Pytorch CUDA Extension & N-MNIST & 92.7 \\
        \textbf{\cite{sparse_spike_gd}} & \textbf{3 layer SNN} & \textbf{Sparse Backpropagation} & \textbf{Pytorch CUDA Extension} & \textbf{N-MNIST} & \textbf{97.4} \\  \hline

        \cite{sparse_spike_gd} & 3 layer SNN & Backpropagation & Pytorch CUDA Extension & F-MNIST & 82.2 \\
        \cite{sparse_spike_gd} & 3 layer SNN & Sparse Backpropagation & Pytorch CUDA Extension & F-MNIST & 80.1 \\ 
        \cite{sparse_spike_gd} & 6 layer SNN & Sparse Backpropagation & Pytorch CUDA Extension & F-MNIST & 82.7 \\
        \textbf{\cite{sparse_spike_gd}} & \textbf{CSNN} & \textbf{Backpropagation} & \textbf{Pytorch CUDA Extension} & \textbf{F-MNIST} & \textbf{86.9} \\
        \cite{sparse_spike_gd} & CSNN & Sparse Backpropagation & Pytorch CUDA Extension & F-MNIST & 86.7 \\     
        \cite{NC_in_SNNs} & 2 layer SNN & Rate Encoding, STDP & Not Listed & F-MNIST & 68.3 \\ 
        \cite{NC_in_SNNs} & 2 layer SNN & TTFS Encoding, STDP & Not Listed & F-MNIST & 71.3 \\ 
        \cite{NC_in_SNNs} & 2 layer SNN & Phase Encoding, STDP & Not Listed & F-MNIST & 71.4 \\ 
        \cite{NC_in_SNNs} & 2 layer SNN & Burst Encoding, STDP & Not Listed & F-MNIST & 71.3 \\ \hline
        
        \cite{Salaj} & CSNN + RSNN & BPTT & Not Listed & DVS & 97.1 \\ 
        \cite{DECOLLE} & CSNN & Surrogate GD & Not Listed & DVS & 97.5 \\ 
        \cite{DECOLLE} & CSNN & Surrogate GD & Not Listed & DVS & 97.8 \\
        \cite{HATS} & SVM & Time Surfaces + SVM & Not Listed & DVS & 95.2 \\
        \cite{She21} & CSNN & STDP & Not Listed & DVS & 96.2 \\
        \cite{She22} & mMND (SNN) & STDP & Not Listed & DVS & 96.6 \\
        \textbf{\cite{She22}} & \textbf{mMND (SNN)} & \textbf{BPTT} & \textbf{Not Listed} & \textbf{DVS} & \textbf{98.0} \\ \hline
 
        \cite{Wade} & 3 layer SNN & Supervised & BRIAN & TI46 & 95.3 \\
        \textbf{\cite{Zhang_DigiLiqStateMachine}} & \textbf{Reservoir SNN} & \textbf{Supervised} & \textbf{BRIAN} & \textbf{TI46} & \textbf{99.8} \\ 
        \cite{Rathi} & 3 layer SNN & Unsupervised STDP & BRIAN & TI46 & 96.0 \\ \hline
        
        \cite{DSpike} & ResNet-19 & DSpike & Not Listed & CIFAR-10 & 94.2 \\ 
        \textbf{\cite{IMLoss}} & \textbf{ResNet-19} & \textbf{IM-Loss, ESG, \& tdBN} & \textbf{Not Listed} & \textbf{CIFAR-10} & \textbf{95.4} \\ 
        \cite{IMLoss} & VGG-16 & IM-Loss, ESG, \& tdBN & Not Listed & CIFAR-10 & 93.8 \\ 
        \cite{IMLoss} & CIFARNet & IM-Loss, ESG, \& tdBN & Not Listed & CIFAR-10 & 92.2 \\
        \cite{HIRE_SNN} & VGG5 & Not Listed & Pytorch & CIFAR-10 & 87.5 \\
        \cite{HIRE_SNN} & ResNet12 & Not Listed & Pytorch & CIFAR-10 & 90.3 \\  \hline

        \cite{HIRE_SNN} & ResNet12 & Not Listed & Pytorch & CIFAR-100 & 58.9 \\ 
        \cite{HIRE_SNN} & VGG11 & Not Listed & Pytorch & CIFAR-100 & 65.1 \\ 
        \textbf{\cite{DSpike}} & \textbf{ResNet-19} & \textbf{DSpike} & \textbf{Not Listed} & \textbf{CIFAR-100} & \textbf{72.2} \\ 
        \cite{IMLoss} & VGG-16 & IM-Loss, ESG, \& tdBN & Not Listed & CIFAR-100 & 70.1 \\ \hline
    \end{tabular}
    \caption{Summary of common architectures, learning methods, frameworks, datasets, and achieved accuracies.}
\end{table}
\end{tiny}

\end{document}